# Compact Value-Function Representations for Qualitative Preferences

**Ronen I. Brafman**
Computer Science Dept.
Ben-Gurion University,
Beer Sheva, Israel 84105

**Carmel Domshlak**
Computer Science Dept.
Cornell University
Ithaca, NY 14853

**Tanya Kogan**
Computer Science Dept.
Ben-Gurion University
Beer Sheva, Israel 84105

## Abstract

We consider the challenge of preference elicitation in systems that help users discover the most desirable item(s) within a given database. Past work on preference elicitation focused on structured models that provide a factored representation of users' preferences. Such models require less information to construct and support efficient reasoning algorithms. This paper makes two substantial contributions to this area: (1) Strong representation theorems for factored value functions. (2) A methodology that utilizes our representation results to address the problem of optimal item selection.

## 1 INTRODUCTION

In recent years, large information sources have become increasingly accessible to users via the web. Retrieving the right information from such sources can be quite challenging. Search engines can help if we have a good idea about what it is that we are looking for. For example, we can locate a concrete book using a few key words. Yet, often, we do not have a particular item in mind, but we know what properties we would like it to satisfy. For instance, when booking a flight, we typically do not have a particular choice in mind, but we do have preferences and constraints on many of its attributes. Similarly, when shopping for a camera, we may not be familiar with various brands and models, but we will have preferences on its properties.

Preference-based item retrieval must be based on effective, user-friendly preference elicitation tools, followed by efficient determination of an optimal choice [9]. The practical problem of preference elicitation has received increasing attention in the past few years (e.g., see [2, 3, 6, 8, 12, 15, 17]). Much of the work in this area considers structured representations of preference information [11] – whether quantitative (i.e., value and utility functions) or qualitative (i.e., orderings of various kinds). The motivation for this development is clear: factored representations are more compact, can be reasoned with more efficiently, and lead to more intuitive queries to the user.

This papers continues this development, providing two substantial contributions. Our first contribution (Section 3) is a new representation theory for generalized additive value functions. We provide conditions under which there exists a particular factored value function representing the user's preference relation. Our representation theorems show that certain partial orders induced by sets of qualitative statements of conditional preference for variable values and conditional relative importance between different variables, can be represented using a compact generalized additive value function. In some respects, these results are stronger than the classical results on additive value functions [13] (although the latter pertain to continuous variables), and more practical, too. The conditions we require are much weaker than those required for an additive representation, and the statements represented are very natural. Formal proofs appear in an online appendix [1].

Our second contribution (Section 4) is a methodology for preference elicitation which utilizes our representation results and works as follows: First, the user provides us with a set of qualitative preference statements. These statements are used (by means of solving a set of linear constraints) to generate a candidate value function whose structure is based on the qualitative information supplied by the user. The existence of such a compact value function is guaranteed by our representation theorem. The user is presented with the top database items according to this value function and either quits or indicates the best presented item. This information induces additional linear constraints on the value function, resulting in a modified value function and a new set of top candidates. This process continues until the best item is identified.

---

[1] Available at www.cs.cornell.edu/ dcarmel/a.ps

Finally, we describe some experiments conducted with a prototype system for online flight reservations which is based on our methodology. In particular, we show that using our methodology the optimal flight configuration is typically identified within a very small number of iterations, and compare these results to the results achieved with a standard technique from multi-attribute value theory. An extensive discussion of this system appears in [14].

## 2 FORMAL BACKGROUND

Our methodology rests on two pillars: A graphical structure called TCP-net [7] used to organize and model preference statements the user provides, and a family of generalized additive value functions that can be viewed as the target of a knowledge compilation process, using which we can easily compare and rank different database items.

### 2.1 TCP-NETS: THE MODEL AND AN EXAMPLE

Our primary aim is to find the most suitable item within a given database $DB$. Items are basically assignments to a set of attributes/variables $\mathbf{X} = \{X_1, \ldots, X_n\}$ (e.g., departure time, airline, etc.) with domains $\mathcal{D}(X_i)$, thus $DB$ can be viewed as a subset of $\Omega = \times \mathcal{D}(X_i)$. To recognize the most suitable item, we need information about the user's preference relation over different variable assignments. Our basic assumption is that there is a relatively compact and sufficiently accurate representation of the user's preferences, much of which can be captured using the two types of natural preference statements, together with a set of preferential independence assumptions that are implicit in the user's specification.

The statements we consider are very intuitive, thus lay users should find them natural and easy to express: (1) *Conditional and unconditional value preferences.* "I prefer British Airways to Air France" is an unconditional value preference over the domain of the `airline` attribute. "I prefer British Airways to Air France in morning flights" is a conditional value preference over the `airline` domain that is conditioned by the value of the `departure-time` attribute; (2) *Conditional and unconditional relative importance of variables.* "`Departure time` time is more important than `airline`" is an unconditional relative importance statement. "`Departure time` is more important than `airline` if I'm flying business class" is a conditional relative importance statement.

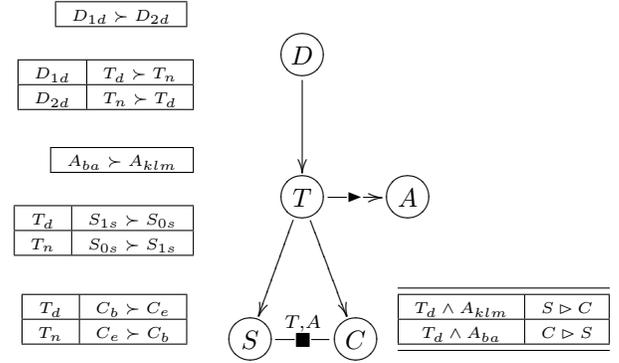

Figure 1: A TCP-net from the flight selection domain.

Provided by the user, preference statements as above can be organized within a graphical structure called TCP-net [7], an extension of the CP-nets model [5]. Formally, a TCP-net $N$ is a tuple $\langle \mathbf{X}, \mathsf{cp}, \mathsf{i}, \mathsf{ci}, \mathsf{cpt}, \mathsf{cit} \rangle$[2]:

1. $\mathbf{X}$ is a set of nodes, corresponding to the problem variables $\{X_1, \ldots, X_n\}$.

2. $\mathsf{cp}$ is a set of directed cp-*arcs* $\langle \overrightarrow{X_i, X_j} \rangle$ s.t. user preferences over the values of $X_j$ depend on the actual value of $X_i$. For each $X \in \mathbf{X}$, let $\mathbf{U}_X = \{X' | \langle \overrightarrow{X', X} \rangle \in N\}$.

3. $\mathsf{i}$ is a set of directed i-*arcs* $(\overrightarrow{X_i, X_j})$. An edge from $X_i$ to $X_j$ denotes $X_i$ is more important than $X_j$. Thus, it is more important to satisfy preference w.r.t. $X_i$ than w.r.t. $X_j$. This is denoted $X_i \rhd X_j$.

4. $\mathsf{ci}$ is a set of undirected ci-*arcs* $(X_i, X_j)$. The existence of such an edge between $X_i$ and $X_j$ implies that there are certain conditions under which one is more important than the other. The relative importance of $X_i$ and $X_j$ is fully determined by the value of some *selector set* $\mathbf{S} \subseteq \mathbf{X} - \{X_i, X_j\}$.

5. $\mathsf{cpt}$ associates with every $X \in \mathbf{X}$ a conditional preference table (CPT), i.e., a mapping from $\mathcal{D}(\mathbf{U}_X)$ to partial orders $\succ_\mathbf{u}$ over $\mathcal{D}(X)$ explicating the actual conditional preferences over $X$.

6. $\mathsf{cit}$ (conditional importance table) associates with every ci-arc $(X_i, X_j)$ a subset $\mathbf{S}$ of $\mathbf{X} - \{X_i, X_j\}$ and a mapping from $\mathcal{D}(\mathbf{S})$ to partial orders $\rhd_\mathbf{s}$ over the set $\{X_i, X_j\}$.

Figure 1 illustrates these ideas by means of an example TCP-net for selecting a business flight from Israel to a conference in USA. This network consists of five variables, standing for various parameters of the flight[3]:

- $D$ – Departure day. Our busy, married user prefers flights leaving one day ($D_{1d}$) before the conference to flights leaving two days ($D_{2d}$) before the conference.

---
[2]A CP-net is simply a TCP-net in which the sets i and ci (and therefore cit) are empty.

[3]Variables in this example are binary, but the semantics of TCP-nets is defined w.r.t. arbitrary finite domains.

- $A$ – Airline. Our user prefers British Airways ($A_{ba}$) to KLM ($A_{klm}$).

- $T$ – Departure time. When leaving two days before the conference, our user prefers the later night flight ($T_n$) to the earlier day flight ($T_d$). When leaving just one day before the conference, these preferences are reversed.

- $S$ – Stop-overs. On day flights, our user would like to have a smoking break, so he prefers an indirect flight ($S_{1s}$) to a direct flight ($S_{0s}$). On night flights, he sleeps well, and so prefers the shorter, direct flight.

- $C$ – Class. On a night flight, our user prefers the cheaper economy ($C_e$) seats because he sleeps anyways, but want to enjoy business class ($C_b$) service on day flights.

CP-arcs and CPTs in Figure 1 capture these preference statements and the underlying preferential dependencies. Our user's relative importance relations are as follows: There is an i-arc from $T$ to $A$: getting a more suitable flying time is more important than getting the preferred airline. There is a ci-arc between $S$ and $C$, where the relative importance of $S$ and $C$ depends on the values of $T$ and $A$ (see the corresponding CIT): (i) On KLM day flights, an intermediate stop in Amsterdam is more important than flying business class. (Our user likes the casino in Amsterdam's airport.) (ii) On British Airways day flights, business class is more important than a stop-over. (Smoking areas in Heathrow are depressing.)

The semantics of a TCP-net $N$ is defined in terms of total preference rankings consistent with the constraints imposed by cpt and cit of $N$, i.e. total rankings consistent with the partial order induced by $N$. The local constraints are interpreted *ceteris paribus*. For example, the fact that in the CPT for departure time ($T$) we have that $T_m \succ T_n$ given $D = D_{1d}$ implies that, given two flights departing one day before the conference that differ *only* in their departure time, the user prefers the one leaving in the morning to the one leaving at night. The fact that $T$ is more important than $A$ implies that given two flights that are identical, except for the value of $T$ and $A$, the user prefers the one in which $T$ is assigned a better value regardless of the value of $A$. Similar semantics is given to conditional importance relation, taking into account the requirement for the conditioning variables (the selector set). A TCP-net $N$ is consistent iff there is some total ranking $\succ$ that is consistent with it. For all $o, o' \in \Omega$, $o \succ o'$ is implied by a TCP-net $N$ (denoted as $N \models o \succ o'$) iff it holds in *all* preference rankings consistent with $N$, and this preferential entailment with respect to a consistent TCP-net is transitive. (For the formal semantics, see [7].)

The structure of the TCP-net is used to (i) recognize the (in)consistency of the statements, (ii) perform efficient inference, and, as we show below, (iii) it can help us identify compactly representable value functions consistent with this preference information. We reemphasize that users are not required to specify the explicit graphical model we just defined, but simply to verbalize statements of the two kinds discussed earlier (which are automatically organized as a TCP-net). Nor do users need to specify the CPTs completely (e.g. as in Bayes nets). This property is especially important in practice as users should not be required to express every nuance of their preferences.

## 2.2 VALUE FUNCTIONS AND GA-DECOMPOSITION

A value function $v : \Omega \rightarrow \mathbb{R}$ is a real-valued function defined over the space of all possible assignments on $\mathbf{X}$ (also referred to as outcomes or items). Value function $v$ is consistent with a (possibly partial) preference ordering $\succeq$ of the user iff $v(o) > v(o')$ for all $o \succ o'$, and $v(o) \geq v(o')$ for all $o \succeq o'$. As the size of $\Omega$ is exponential in the number of problem variables, only compactly representable value functions can be practically useful. In this paper we focus on one such family of compactly representable value functions, namely *generalized additive* (GA) value functions. The notion of GA value functions closely corresponds to the notion of *generalized additive independence* for utility functions [1], but addresses only the structural assumptions of the latter: A value function $v$ over the variables $\mathbf{X}$ is GA if there exists a not necessarily disjoint partition $\mathbf{P}_1, \ldots, \mathbf{P}_k$ of $\mathbf{X}$, such that $v(\mathbf{X}) = \sum_{i=1}^{k} v_i(\mathbf{P}_i)$. In what follows, we refer to these variable subsets $\mathbf{P}_1, \ldots, \mathbf{P}_k$ as the factors of $v$. In fact, *any* value function can be seen as GA (for $k = 1$), but working with the GA-decomposition of a value function is practically feasible only if its factors are sufficiently compact, e.g., the size of each $\mathbf{P}_i$ is bounded by a constant.

## 3 GA-DECOMPOSITION OF TCP-NETS

Value functions provide a mathematically general and efficient way of representing and reasoning with preference information. Given a value function, we can quickly sort a database or determine the top-$k$ of its items. However, obtaining a value function directly from the user is significantly more involved than obtaining a set of simple preference statements.

Therefore, we propose using a TCP-net to initially organize *qualitative* preference statements obtained from the user and their dependency structure; compile this information to a value function that *main-* tains the qualitative structure and independence assumptions implicit in this TCP-net; use it to sort the database items; and later, possibly refine this value function using additional qualitative information (see Section 4), while still maintaining independence assumptions implied by the original TCP-net. Thus, the main theoretical question we face is: *Given a TCP-net $N$ as input, can we efficiently generate a value function $v$ that is consistent with $N$?* Of course, it is quite trivial to see that, for any consistent TCP-net, there exists at least one value function that represents it. Indeed, this is true for any partial order. However, what we would really like to know is whether we can find a *structured* value function that, in some sense, is *as compact as the original TCP-net*. Specifically, we would like to know whether there exists a GA value function defined over small factors "implied" by the structure of the network.

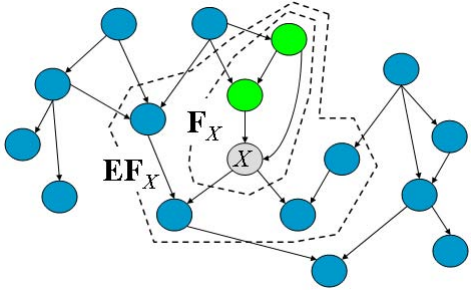

Figure 2: CP-family and extended CP-family of $X$.

To answer this question, we consider three progressively more complicated classes of TCP-nets, and show how the factors of GA value functions representing these TCP-nets relate to their graphical structure. We prove this relation by showing that for *every* TCP-net in each of these network classes there exists a GA value function over a particular set of factors. Likewise, our representation results provide us with a concrete computational mechanism for generating such GA value functions, tractable for a wide class of TCP-nets.

### 3.1 CP-NETS

First, we consider CP-nets (i.e., TCP-nets with only cp-arcs), and start with some notation. Given a CP-net $N$ over variables $\mathbf{X}$, let $\mathbf{U}_X$ and $\mathbf{Y}_X$ be the sets of immediate ancestors and immediate descendants of $X$ in $N$, respectively. Let $\mathbf{F}_X = \{X\} \cup \mathbf{U}_X$ denote the *CP-family* of $X$, and $\mathbf{EF}_X = \mathbf{F}_X \cup \bigcup_{Y \in \mathbf{Y}_X} \mathbf{F}_Y$ denote the *extended CP-family* of $X$ (see Figure 2). Now we can specify the CP-condition that plays a central role in our discussion. Originally, CP-conditions were introduced in [4] for defining real-valued value functions representable as UCP-nets.

**Definition 1** Given a CP-net $N$, and a set of functions $\Phi = \{\phi^{X_1}, \ldots, \phi^{X_n}\}$ over $\mathbf{F}_{X_1}, \cdots, \mathbf{F}_{X_n}$, respectively, we say that $\Phi$ satisfies the *CP-conditions* of $N$ if and only if for each variable $X \in \mathbf{X}$, each $\mathbf{u} \in \mathcal{D}(\mathbf{U}_X)$, and each $x_1, x_2 \in \mathcal{D}(X)$, if $N \models x_1 \succ_{\mathbf{u}} x_2$, then for each $\mathbf{v} \in \mathcal{D}(\mathbf{EF}_X - \{X\})$ compatible with $\mathbf{u}$ (i.e. $\mathbf{v}$ and $\mathbf{u}$ provide the same assignment to the shared variables), we have:

$$\phi^X(x_1, \mathbf{u}) + \sum_{i=1}^{|\mathbf{Y}_X|} \phi^{Y_i}(\mathbf{v}_i, x_1) > \phi^X(x_2, \mathbf{u}) + \sum_{i=1}^{|\mathbf{Y}_X|} \phi^{Y_i}(\mathbf{v}_i, x_2) \quad (1)$$

where $\mathbf{v}_i$ is the value provided by $\mathbf{v}$ to $(\mathbf{F}_{Y_i} - \{X\})$.

Developed by Boutilier *et al.* [4], Lemma 1 below exploits the CP-conditions of $N$ to provide a necessary and sufficient condition for a GA value function with factors $\mathbf{F}_{X_1}, \cdots, \mathbf{F}_{X_n}$ to be consistent with $N$.

**Lemma 1 ([4])** *Given a CP-net $N$, and a function*

$$v(\mathbf{X}) = \sum_{i=1}^{n} \phi^{X_i}(\mathbf{F}_{X_i}), \quad (2)$$

*we have $v$ consistent with $N$ iff $\{\phi^{X_1}, \ldots, \phi^{X_n}\}$ satisfy the CP-conditions of $N$.*

In fact, Lemma 1 provides us with even stronger knowledge on GA-decomposability of $N$. First, reasonably assuming that no preferential dependency between $X$ and $\mathbf{U}_X$ is redundant, one could not expect a more compact GA-decomposition of $N$. Second, CP-conditions actually provide us with a concrete procedure for generating such a value function $v$:

(1) Given a CP-net $N$, construct a system of linear inequalities $L$, variables of which stand for the entries of the sub-functions $\phi^{X_1}, \ldots, \phi^{X_n}$ and inequalities correspond to all the required instances of Eq. 1. Let $\mathcal{H}_L$ be the polytop defined by $L$.

(2) If $L$ is satisfiable (i.e. $\mathcal{H}_L$ is not empty), pick *any* solution for $L$ (either by solving a linear program defined by $L$ and an arbitrary objective function bounded on $\mathcal{H}_L$, or by sampling a point from $\mathcal{H}_L$).

Step (2) is correct because *any solution for $L$ constitutes a value function $v$ of form (2), consistent with $N$.* The complexity of $L$ is locally exponential: the number of variables and inequalities in $L$ is $O(nd^\lambda)$ and $O(nd^{2\mu})$, respectively, where $d = \max_{X \in \mathbf{X}} \{|\mathcal{D}(X)|\}$, $\lambda = \max_{X \in \mathbf{X}} \{|\mathbf{F}_X|\}$, and $\mu = \max_{X \in \mathbf{X}} \{|\mathbf{EF}_X|\}$. Finally, since linear programming is in P, we obtain the following corollary of practical interest.

**Corollary 2** *If a CP-net $N$ is GA-decomposable over its CP-families, and we have $\max_{X \in \mathbf{X}} \{|\mathbf{EF}_X|\} = k$ for some constant $k$, then a value function $v(\mathbf{X}) = \sum_{i=1}^{n} \phi^{X_i}(\mathbf{F}_{X_i})$ consistent with $N$ can be constructed in time polynomial in the size of $N$.*

Corollary 2 presents a wide class of efficiently GA-decomposable CP-nets. However, notice that nothing so far prevents $\mathcal{H}_L$ from being empty, since Lemma 1 provides no guarantees for the actual GA-decomposability. It is possible that, for some CP-nets, value functions of form (2) simply do not exist. As we would like to assume that user's statements provide us with sufficient information about value independence, such incompleteness would clearly be problematic. Fortunately, Theorem 3 below shows that polytops $\mathcal{H}_L$ for *acyclic CP-nets* are always non-empty[4].

**Theorem 3** *Every acyclic CP-net is GA-decomposable over its CP-families.*

To relate Theorem 3 to the classical results in this area, consider a CP-net without any edges. According to Theorem 3, such a CP-net induces an additive value function. Indeed, variables in such a CP-net are mutually preferentially independent, a necessary and sufficient condition for additive decomposition (see Theorem 3.6 in [13]). Thus, a representation theorem for standard additive value functions is a special case of our Theorem 3. As far as we know, results on conditional structures and generalized additive decomposability exist for *utility* functions [1], but require complex conditions which do not seem to relate in any simple manner to the above result. We note that the classical results on additive decomposition refer to continuous variables, whereas we deal with discrete variables, only.

### 3.2 TCP-NETS WITH NO ci-ARCS

Now, let us consider a wider class of TCP-nets, namely *TCP-nets with no ci-arcs*. Here we show that GA-decomposability of this class of networks deserves special attention. We begin by formalizing a new set of conditions essential for analysis of GA-decomposability of this class of networks.

**Definition 2** Given a TCP-net $N$ with no ci-arcs, and a set of functions $\Phi = \{\phi^{X_1}, \ldots, \phi^{X_n}\}$ over $\mathbf{F}_{X_1}, \cdots, \mathbf{F}_{X_n}$, respectively, we say that $\Phi$ satisfies the *I-conditions* of $N$ if and only if for each i-arc $\overrightarrow{(X, X')} \in N$, each $\mathbf{u} \in \mathcal{D}(\mathbf{U}_X)$, and each $x_1, x_2 \in \mathcal{D}(X)$, if $N \models x_1 \succ_{\mathbf{u}} x_2$ then, for each $x_1', x_2' \in \mathcal{D}(X')$, each $\mathbf{u}' \in \mathcal{D}(\mathbf{U}_{X'})$ compatible with $\mathbf{u}$, and each $\mathbf{v} \in \mathcal{D}(\mathbf{EF}_X - \{X, X'\})$, $\mathbf{v}' \in \mathcal{D}(\mathbf{EF}_{X'} - \{X, X'\})$ compatible with $\mathbf{u}$ and $\mathbf{u}'$, we have:

$$\phi^X(x_1, \mathbf{u}) + \phi^{X'}(x_1', \mathbf{u}') + \sum_{i=1}^{|\mathbf{Y}_X|} \phi^{Y_i}(\mathbf{v}_i, x_1, x_1') + \sum_{i=1}^{|\mathbf{Y}_{X'} - \mathbf{Y}_X|} \phi^{Y_i'}(\mathbf{v}_i', x_1', x_1) >$$
$$\phi^X(x_2, \mathbf{u}) + \phi^{X'}(x_2', \mathbf{u}') + \sum_{i=1}^{|\mathbf{Y}_X|} \phi^{Y_i}(\mathbf{v}_i, x_2, x_2') + \sum_{i=1}^{|\mathbf{Y}_{X'} - \mathbf{Y}_X|} \phi^{Y_i'}(\mathbf{v}_i', x_2', x_2)$$
(3)

where $\mathbf{v}_i$ and $\mathbf{v}_i'$ are the values provided by $\mathbf{v}$ and $\mathbf{v}'$ to $(\mathbf{F}_{Y_i} - \{X, X'\})$ and $(\mathbf{F}_{Y_i'} - \{X, X'\})$, respectively. Note that $x_1'$ and $x_2'$ (similarly, $x_1$ and $x_2$) might be redundant parameters for some $\phi^{Y_i}$ (respectively, $\phi^{Y_i'}$).

It turns out that the I and CP-conditions together constitute for TCP-nets with no ci-arcs *exactly* what the CP-conditions alone constitute for CP-nets.

**Lemma 4** *Given a TCP-net $N$ with no ci-arcs, and a function $v$ of form (2), we have $v$ consistent with $N$ iff $\{\phi^{X_1}, \ldots, \phi^{X_n}\}$ satisfy CP- and I-conditions of $N$.*

Notice that, similarly to the case of CP-nets, Lemma 4 provides TCP-nets with no ci-arcs with a procedure for generating consistent GA value functions of form (2). However, an immediate concern should be about its usefulness: At first sight, such decomposition does not seem to be very likely, as the functional form (2) is based only on preference dependencies, completely ignoring the importance relations induced by the i-arcs. Theorem 5 shows that these concerns are not entirely justified, and that value decomposition of form (2) is complete for *acyclic TCP-nets with no ci-arcs*. (Since all the arcs in such networks are directed, the corresponding notion of acyclicity is straightforward.)

**Theorem 5** *Every acyclic TCP-net with no ci-arcs is GA-decomposable over its CP-families.*

Theorem 5 shows that additional unconditional relative importance relation do not affect GA-decomposability of the network (assuming it remains consistent). Lemma 4 shows that any such GA value function corresponds to a solution of a linear system $L$, as in the case of CP-nets. Still locally exponential, the complexity of $L$, however, is affected by i-arcs, since now $L$ consists of both instances of Eq. 1 and Eq. 3. As a result, the number of variables in $L$ is still $O(nd^\lambda)$, but the number of equations grows to $O((n + l)d^{2\mu})$, where $l$ is the number of i-arcs in $N$. Notice that the order of description complexity of $L$ remains the same as for CP-nets, thus Corollary 2 can be re-stated for TCP-nets with no ci-arcs, all else being equal.

### 3.3 TCP-NETS

Let us now consider TCP-nets capturing all our types of preference statements, thus consisting of both di-rected (cp and i) and undirected (ci) arcs. Let $\mathbf{S}_X$ be the union of selector sets of ci-arcs involving $X$ (i.e., $\mathbf{S}_X = \bigcup_{\gamma=(X,X')} \mathbf{S}_\gamma$), and, reversely, let $\mathbf{W}_X$ be the set of end-point variables of ci-arcs where $X$ acts as a selector (i.e. $\mathbf{W}_X = \{X' \mid X \in \mathbf{S}_{X'}\}$). Let $\mathbf{Y}^\star_X = \mathbf{Y}_X \cup \mathbf{W}_X$ be the set of all $X$'s dependents. Let $\mathbf{I}_{X|\mathbf{s}}$ be the set of all variables $X'$ that are directly more important than $X$ given $\mathbf{s} \in \mathcal{D}(\mathbf{S}_X)$. Finally, let $\mathbf{F}^\star_X = \mathbf{F}_X \cup \mathbf{S}_X$ denote the *TCP-family* of $X$ (see Figure 3, where the dashed arcs *schematically* connect between $\mathbf{S}_X$ and $X$), and $\mathbf{EF}^\star_X = \mathbf{F}^\star_X \bigcup_{Y \in \mathbf{Y}^\star_X} \mathbf{F}^\star_Y$ denote the *extended TCP-family* of $X$.

---
[4]The proof of Theorem 3 is based on a technique developed in [10] for compiling acyclic CP-nets into systems of soft constraints.

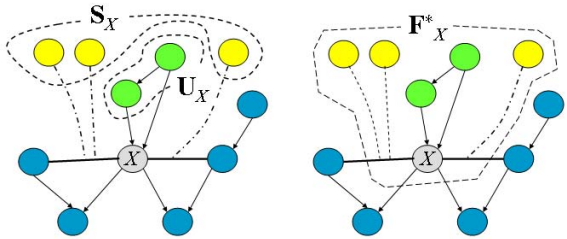

Figure 3: TCP-family of $X$.

TCP-nets with ci-arcs are significantly richer than these without, and GA-decomposition of form (2) is not expressive enough to cover these type of networks. However, here we show that there exists a sufficiently expressive (yet often compact) extended counterpart of Eq. 2, namely:

$$v(\mathbf{X}) = \sum_{i=1}^n \phi^{X_i}(\mathbf{F}^\star_{X_i}). \qquad (4)$$

**Definition 3** Consider a TCP-net $N$, and a set of functions $\Phi = \{\phi^{X_1}, \ldots, \phi^{X_n}\}$ over $\mathbf{F}^\star_{X_1}, \cdots, \mathbf{F}^\star_{X_n}$, respectively. We say that $\Phi$ satisfies the *CI-conditions* of $N$ if and only if for each ci-arc $(X, X') \in N$ (and, similarly, each i-arc $\overrightarrow{(X, X')} \in N$), each $x_1, x_2 \in \mathcal{D}(X)$, and each $\mathbf{u} \in \mathcal{D}(\mathbf{U}_X)$ and $\mathbf{s} \in \mathcal{D}(\mathbf{S}_X)$, if $X \in \mathbf{I}_{X'|\mathbf{s}}$ and $N \models x_1 \succ_\mathbf{u} x_2$, then, for each $x'_1, x'_2 \in \mathcal{D}(X')$, and each set of (all compatible) $\mathbf{u}' \in \mathcal{D}(\mathbf{U}_{X'})$, $\mathbf{s}' \in \mathcal{D}(\mathbf{S}_{X'})$, $\mathbf{v} \in \mathcal{D}(\mathbf{EF}^\star_X - \{X, X'\})$, $\mathbf{v}' \in \mathcal{D}(\mathbf{EF}^\star_{X'} - \{X, X'\})$, we have:

$$\begin{aligned}
&\phi^X(x_1, \mathbf{u}, \mathbf{s}) + \phi^{X'}(x'_1, \mathbf{u}', \mathbf{s}') + \\
&\sum_{i=1}^{|\mathbf{Y}^\star_X|} \phi^{Y_i}(\mathbf{v}_i, x_1, x'_1) + \sum_{i=1}^{|\mathbf{Y}^\star_{X'} - \mathbf{Y}^\star_X|} \phi^{Y'_i}(\mathbf{v}'_i, x'_1, x_1) > \\
&\phi^X(x_2, \mathbf{u}, \mathbf{s}) + \phi^{X'}(x'_2, \mathbf{u}', \mathbf{s}') + \\
&\sum_{i=1}^{|\mathbf{Y}^\star_X|} \phi^{Y_i}(\mathbf{v}_i, x_2, x'_2) + \sum_{i=1}^{|\mathbf{Y}^\star_{X'} - \mathbf{Y}^\star_X|} \phi^{Y'_i}(\mathbf{v}'_i, x'_2, x_2)
\end{aligned} \qquad (5)$$

Lemma 6 below shows that CI and CP-conditions[5] are necessary and sufficient for GA-decomposability of general TCP-nets along the functional form (4).

---

[5]To fit the functional form 4, the CP-conditions should be simply reformulated from CP- to TCP-families.

**Lemma 6** *Given a TCP-net $N$, and a function $v$ of form (4), we have $v$ consistent with $N$ iff $\{\phi^{X_1}, \ldots, \phi^{X_n}\}$ satisfy CP- and CI-conditions of $N$.*

It is unlikely that every consistent TCP-net is GA-decomposable along the functional form (4). Yet, we showed that such decomposability is complete for *acyclic TCP-nets*. Since TCP-nets may contain both directed and undirected arcs, the corresponding notion of acyclicity is non-standard: A TCP-net $N$ is *acyclic* if each cycle in its induced undirected graph, when projected back to $N$, contains directed arcs in different directions.

**Theorem 7**
*Every acyclic TCP-net is GA-decomposable over its TCP-families.*

Theorem 7 finalizes our representation theory. Clearly, adding ci-arcs reduce the general compactness of GA-decomposition, but factoring on TCP- instead of CP-families seem to be unavoidable.

## 4  METHODOLOGY AND EXPERIMENTS

Our representation results for conditional preference statements provide us with formal ground for a novel methodology for preference-based item retrieval in which *qualitative preference elicitation is followed by quantitative reasoning with preferences.*

**Step 1:** *Obtain initial preferences.* The user is allowed to express her preference and importance statements (subsequently organized as an acyclic TCP-net $N$), as well as her hard constraints on various flight parameters. Using the same module, we can also simulate diverse classes of users, controlling various TCP-net properties such as structure, denseness of CPTs, local structure in CPTs, etc.

**Step 2:** *Generate value function.* The system generates and solves the system of linear constraints $L$ that $N$ poses on the space of compact GA value functions. Depending on the structure of $N$, $L$ corresponds to one of the linear systems described in Section 3. Thus, if $L$ is satisfied, the GA value function $v$ corresponding to $L$'s solution is consistent with $N$. Otherwise, the original specification of $N$ has to be inconsistent, and this inconsistency has to be resolved by the user.

**Step 3:** *Display-and-Feedback.* The user is presented with the top $k$ database items according to $v$ ($k = 10$ in our system). If the user is satisfied with one of these $k$ items, we are done. Otherwise, she indicates the most preferred item within the list. The practi-cal rational behind this step is that users are used to similar interactions with search engines.

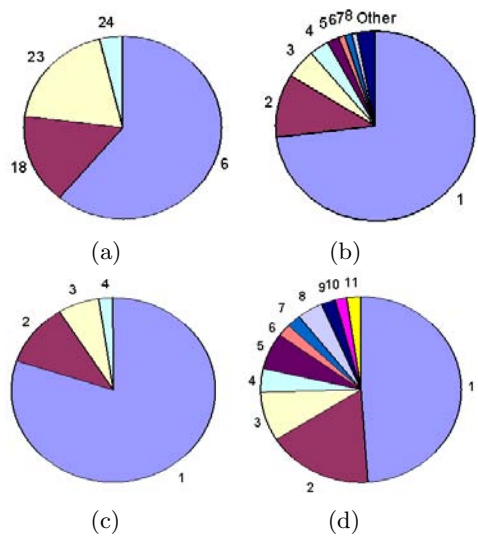

Figure 4: Experimental results.

**Step 4:** *Update-and-Repeat.* The new information obtained from the user provides $k-1$ additional linear constraints (representing the fact that the selected item is better than any of the other $k-1$ items), and these constraints are added to $L$. Now, we return to Step 2 and repeat the process.

The idea of generate-and-revise and the use of linear constraints to generate value functions is not new. The most notable example is the VEIL system [2]. What distinguishes our system from VEIL is: (1) the use of a rich qualitative input language; (2) the ability to model conditional preferences and importance relations, and the corresponding use of GA value functions; (3) the use of top-$k$ selection, motivated by search engines. VEIL, on the other hand, is based on additive value functions that cannot represent conditional preferences. The user must input a fixed value function for each attribute. The system's task is basically to learn the relative weight of each of the (attribute) value functions. Finally, to the best of our knowledge, the only previous work considering LP compilation of a set of qualitative preference statements into a value function is [16]. In comparison to [16], our graphical analysis of conditional preferential independence between the variables leads to representation theorems providing significantly more compact and more efficient value-function decomposition.

To evaluate the utility of our approach, we implemented a test-bed system in which users can perform flight reservations, and run a set of experiments with "simulated" users. This system acts as a proxy between the user and an on-line database of Orbitz.com.

Once the user's TCP-net and hard constraints are available, a standard query reflecting only the user's hard constraints is automatically generated and posed to the on-line database. The result of this query is stored and constitutes our actual database.

We tested our system with real data of New-York/Boston flights using virtual users. Each "user" is simulated with a random GA value function $v$ that is (1) consistent with a random acyclic TCP-net representing user's input, and (2) composed of the factors as in Section 3. Note that (2) assumes that the dependencies in user's preferences are completely captured by the TCP-net. Based on $v$, we now simulate the user/system interaction by repeatedly selecting the top item among the 10 currently displayed until the system agrees with the "user" on the top item. In each such experiment, we measure the number of rounds (i.e., value-function updates based on user feedback) required to recognize this top item. Note that this corresponds to a somewhat worst-case evaluation settings, since in real-life the user might be satisfied by the best item of one of the intermediate iterations. Below we present a small sample of our results, referring the reader to [14] for a comprehensive description.

Fist, we tried to evaluate the utility of assuming a generalized additive model rather than the basic additive model used in past work. We compared the number of iterations required to recognize the best element using generalized additive (Figure 4b) as opposed to additive (Figure 4a) structures when the true user preferences are conditional. We see that the additive models perform poorly, i.e. when conditional structures are not modeled, many more iterations are required. More importantly, we see that our methodology is able to recognize the top element immediately in 73% of the cases, and in up to 3 iterations in 89% of the cases. This indicates that the standard technique in multi-attribte value theory of using additive value functions to approximately model users with conditional preferences does not work too well.

Figure 4(c-d) depicts the effect of missing information on the system's performance. Figure 4c shows the distribution of number of iterations to convergence when the CPTs are completely specified and values are totally ordered. In contrast, Figure 4d depicts the same distribution but for partially specified CPTs with partially ordered values. As we would expect, the stronger our initial information, the faster we converge on the optimal item. However, Figure 4d shows that even with very partial information about the detailed preferences, knowing the qualitative structure of conditional preferential independence allowed in 76% of the cases to converge within just 3 iterations.

## 5 CONCLUSION

We presented a new representation theory for factored value functions that allow for more useful structures than those appearing in classical textbook results in this area [13]. Our representation theorem shows that certain partial orders, induced by sets of qualitative statements of conditional preference and conditional relative importance, can be represented using compact generalized additive value functions. Exploiting these representation results, we presented a novel methodology for preference-based retrieval of database items. This methodology is based on the use of natural qualitative preference statements, comparisons between top-$k$ elements, and compilation of this information into a factored value function. To the best of our knowledge, this is the most expressive formally-grounded methodology for preference-based item retrieval today. A prototype system based on this methodology was implemented in the domain of flight reservations, and numerous experiments were conducted to understand the behavior of this system.

Our work raises numerous open theoretical questions: Are there compact GA value functions for every consistent (possibly cyclic) TCP-net? Can we characterize the representation theorem purely in terms of conditional independence, without using the graph structure explicitly, or, alternatively, what are the core properties of the graph that allow for a compact GA decomposition? We hope to address these issues in the future.

**Acknowledgements**

We are grateful to the anonymous referees for their useful comments. R. Brafman was partially supported by the Paul Ivanier Center for Robotics and Production Management. C. Domshlak was partially supported by the Intelligent Information Systems Institute, Cornell Univ. (AFOSR grant F49620-01-1-0076).